%% file: main.tex
\documentclass[10pt,conference]{IEEEtran}
\usepackage[colorlinks,urlcolor=blue,linkcolor=blue,citecolor=blue]{hyperref}
\usepackage{caption}
\usepackage{color,array}

\usepackage{graphicx}
\usepackage{listings}
\usepackage{xspace}
\usepackage{listings}
\usepackage{color}
\usepackage{colortbl}
\usepackage[table]{xcolor}
\usepackage{multirow}
\usepackage{amsmath}
\usepackage{hhline}
\usepackage{bigstrut}
\usepackage{rotating}
\usepackage{tcolorbox}
\usepackage{float}
\usepackage{tabularx}
\usepackage{makecell}
\usepackage{booktabs} 
\usepackage{url}
\usepackage{xcolor}
\usepackage{amsmath,amssymb,amsfonts}

\usepackage[table]{xcolor}
\definecolor{ffe1da}{RGB}{255,225,218}
\definecolor{F7E0D5}{RGB}{247,224,213}
\definecolor{darkF7E0D5}{RGB}{209,154,128}
\definecolor{Light}{RGB}{247,224,213}
\usepackage{listings}

\newcommand{\tecoafour}{TeCoA\textsuperscript{4}\@\xspace}
\newcommand{\tecoatwo}{TeCoA\textsuperscript{2}\@\xspace}
\newcommand{\farefour}{FARE\textsuperscript{4}\@\xspace}
\newcommand{\simcliptwo}{Sim-CLIP\textsuperscript{2}\@\xspace}
\newcommand{\simclipfour}{Sim-CLIP\textsuperscript{4}\@\xspace}
\newcommand{\faretwo}{FARE\textsuperscript{2}\@\xspace}
\newcommand{\epsilontwo}{$\epsilon= \nicefrac{2}{255}$\xspace}
\newcommand{\epsilonfour}{$\epsilon=\nicefrac{4}{255}$\xspace}
\newcommand{\epsiloneight}{$\epsilon=\nicefrac{8}{255}$\xspace}
\newcommand{\elfinity}{$\ell_\infty$\xspace}
\usepackage{algorithm}
\usepackage{nicefrac}
\definecolor{missing}{HTML}{BDFCBE}
\definecolor{wrong}{HTML}{F8A6A6}
\definecolor{correct}{HTML}{7CF97E}

\definecolor{mistakes}{HTML}{FFC02B}
\floatstyle{ruled}
\newfloat{listing}{tb}{lst}{}
\floatname{listing}{Listing}
\usepackage{newfloat}
\DeclareCaptionStyle{ruled}{labelfont=normalfont,labelsep=colon,strut=off} 
\floatstyle{ruled}
\newfloat{listing}{tb}{lst}{}
\floatname{listing}{Listing}

\begin{document}

\title{Sim-CLIP: Unsupervised Siamese Adversarial Fine-Tuning for Robust and Semantically-Rich Vision-Language Models}


\author{
Md Zarif Hossain \quad
Ahmed Imteaj \quad\\
Florida Atlantic University \quad \\
\vspace{0.2em}
\texttt{\small\{mdzarifhossa2025, aimteaj\}@fau.edu}\\
\texttt{\url{https://speedlab-git.github.io/Sim-CLIP/}}\\
}

\maketitle

\begin{abstract}
Vision–Language Models (VLMs) rely heavily on pretrained vision encoders to support downstream tasks such as image captioning, visual question answering, and zero-shot classification. Despite their strong performance, these encoders remain highly vulnerable to imperceptible adversarial perturbations, which can severely degrade both robustness and semantic quality in multimodal reasoning.
In this work, we introduce Sim-CLIP, an unsupervised adversarial fine-tuning framework that enhances the robustness of the CLIP vision encoder while preserving overall semantic representations. Sim-CLIP adopts a Siamese training architecture with a cosine similarity objective and a symmetric stop-gradient mechanism to enforce semantic alignment between clean and adversarial views. This design avoids large-batch contrastive learning and additional momentum encoders, enabling robust training with low computational overhead.
We evaluate Sim-CLIP across multiple Vision–Language Models and tasks under both targeted and untargeted adversarial attacks. Experimental results demonstrate that Sim-CLIP consistently outperforms state-of-the-art robust CLIP variants, achieving stronger adversarial robustness while maintaining or improving semantic fidelity. These findings highlight the limitations of existing adversarial defenses and establish Sim-CLIP as an effective and scalable solution for robust vision–language representation learning.
\end{abstract}




\vspace{-0.2cm}

\input{Introduction}
\input{RelatedWorks}

\input{Methodology}
\input{Experimental}

\input{Conclusion}

\bibliographystyle{ieeetr}
\bibliography{ref}

\end{document}

%% file: Introduction.tex
\section{Introduction}
\IEEEPARstart{T}{he} success of Large Language Models (LLMs) \cite{openai2023gpt, meta2023introducing} in text understanding and generation has motivated the extension of their capabilities to vision-centric applications in consumer technologies, including smartphones, wearables, smart assistants, and connected cameras. This progression has given rise to Vision–Language Models (VLMs), which are designed to jointly model and reason over visual and textual inputs. To extract rich visual representations, VLMs typically rely on pretrained vision encoders such as CLIP \cite{radford2021learning}, BEiT \cite{bao2021beit}, and DINO \cite{caron2021emerging}. Leveraging these pretrained encoders allows VLMs to inherit extensive visual knowledge from large-scale multimodal datasets \cite{lin2014microsoft, marino2019ok}, thereby improving performance on downstream tasks without the need for extensive task-specific fine-tuning. Among existing vision encoders, CLIP \cite{radford2021learning} has attracted particular attention and serves as a backbone for many state-of-the-art VLMs, including OpenFlamingo \cite{awadalla2023openflamingo} and LLaVA \cite{liu2024visual}. While the widespread availability of pretrained VLM architectures and weights has accelerated research and real-world adoption, it has also introduced significant safety challenges that demand careful consideration.
Recent studies \cite{zhao2024evaluating, wei2024jailbroken} demonstrate that VLMs are highly vulnerable to adversarial manipulations targeting both textual and visual modalities. Notably, the visual modality has been shown to be particularly susceptible to adversarial perturbations relative to text-based inputs \cite{goodfellow2014explaining, carlini2024aligned}, raising serious concerns for safety-critical and consumer-facing applications.
Besides, the authors in \cite{schlarmann2023adversarial} demonstrate that adversaries can employ human-imperceptible adversarial perturbations to create targeted attacks, effectively generating desired malicious outputs.
 
Although prior methods \cite{mao2022understanding, schlarmann2024robust} have improved the adversarial robustness of CLIP models, substantial challenges persist. In particular, adversarial fine-tuning often incurs a noticeable degradation in downstream task performance. For example, in image captioning, robust CLIP variants frequently fail to preserve global semantic coherence when processing perturbed inputs. This degradation limits their ability to generate accurate and contextually rich descriptions, thereby reducing their effectiveness in real-world consumer applications such as automated doorbells and smart cameras.
To address these limitations, we propose Sim-CLIP, an unsupervised adversarial fine-tuning framework that strengthens the CLIP vision encoder against adversarial attacks while preserving semantic fidelity. Sim-CLIP maintains both local visual detail and global semantic understanding, enabling Vision–Language Models to produce coherent and semantically precise outputs even under targeted perturbations.
A key advantage of Sim-CLIP is its plug-and-play design: it integrates seamlessly into existing VLM pipelines and supports a wide range of downstream tasks without requiring task-specific retraining, a critical property for scalable deployment in consumer-facing systems.

%% file: RelatedWorks.tex
\section{Related Works}

\begin{figure*}[t]
\setlength{\belowcaptionskip}{-15pt}
    \centering
    \includegraphics[width = 0.92\textwidth]{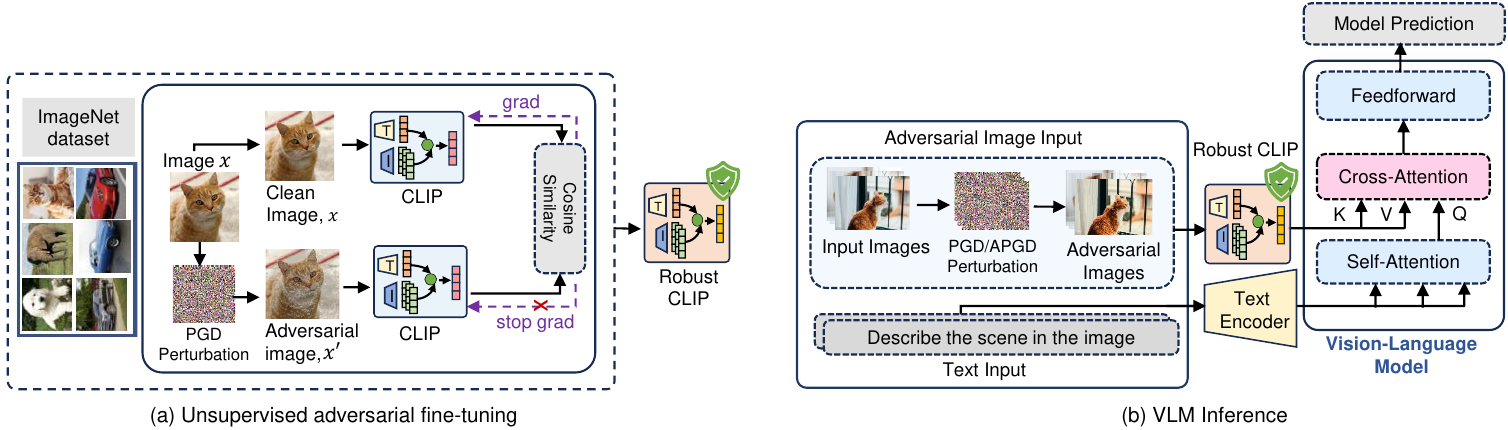}
    \caption{
\textbf{Sim-CLIP workflow:}
(a) CLIP is adversarially fine-tuned with Sim-CLIP and used as the VLM vision encoder.
(b) The robust encoder processes adversarial images with text via cross-attention to generate predictions.
}
    \label{fig:llm}
\end{figure*}

\subsubsection{\textbf{Adversarial robustness in traditional ML}}
The susceptibility of traditional machine learning models, such as CNNs and RNNs, to adversarial attacks has been extensively studied. Prior works has predominantly focused on monomodal settings, targeting either visual or textual modalities in isolation \cite{zheng2024unified}. In the visual domain, gradient-based attacks \cite{wang2024transferable} introduce carefully crafted imperceptible perturbations to input images, while patch-based attacks \cite{zhou2024pair} apply localized adversarial patches to mislead models without modifying the entire image.
Text-based adversarial attacks \cite{ebrahimi2017hotflip, guo2021gradient} have also been widely studied, but require different attack strategies due to the discrete nature of language. To defend against these threats, adversarial training \cite{pan2022improved} has emerged as an effective paradigm.

\vspace{4pt}
\subsubsection{\textbf{Adversarial robustness for VLMs}}
Few recent studies \cite{bansal2023cleanclip,zhou2023advclip,li2024language} demonstrate CLIP's vulnerability to imperceptible attacks, which can significantly impact downstream task performance of VLMs. For instance, in AdvCLIP \cite{zhou2023advclip}, the authors generate universal patches for CLIP models that can deceive VLMs across all of their downstream tasks. In \cite{zhao2024evaluating}, the authors leverage diffusion models to create adversarial samples that manipulate the model into generating a targeted output. Moreover, in \cite{schlarmann2023adversarial}, the authors demonstrate the potential of gradient-based attacks on VLMs, compelling the model to generate inaccurate results.
One recent study \cite{mao2022understanding} introduced a supervised adversarial fine-tuning scheme for CLIP that employs cross-modal image-text contrastive loss. A few concurrent works \cite{jiang2020robust, fan2021does} proposed unsupervised adversarial training methods based on SimCLR contrastive loss. However, these methods necessitate a large batch size to achieve strong robust performance. The authors in \cite{gowal2020self} proposed an adversarial fine-tuning scheme based on BYOL \cite{grill2020bootstrap}, which addresses the issue of large batch sizes but introduces an overhead with the momentum encoder. Besides, the authors in \cite{schlarmann2024robust} proposed an $\ell_2$ loss-based unsupervised fine-tuning scheme, but it fails to capture semantic features and nuanced details of the images effectively. In contrast, our unsupervised fine-tuning approach, based on Siamese architecture, utilizes cosine similarity to effectively capture semantic information during adversarial training without requiring a large batch size or an additional momentum encoder.

%% file: Methodology.tex
\section{Methodology}
\subsection{\textbf{Unsupervised Adversarial Fine-Tuning}} 
Advancing the robustness of VLMs has been a central focus of recent research, particularly in addressing their vulnerability to adversarial attacks. A notable effort in this direction is FARE \cite{schlarmann2024robust}, which introduced an unsupervised adversarial fine-tuning scheme for the CLIP vision encoder that does not rely on text embedding. FARE achieves its robustness by minimizing a $\ell_2$ loss function between an adversarially perturbed image embedding and a clean image embedding. The embedding loss can be expressed as follows: 
\begin{equation}
\mathcal{L}_F (x,x_a) =\max _{\left\|x_a-x\right\|_{\infty} \leq \varepsilon} \| \theta(x_a) - \theta(x) \|_2^2
\end{equation}

\vspace{-0.1cm}
\begin{table*}[htb!]
\small
\centering
\tabcolsep=2.8pt
\extrarowheight=-0.99pt
\caption{Robustness comparison of VLMs under untargeted attacks:
Image captioning performance is evaluated on COCO and Flickr30k using CIDEr, while VQA performance is assessed on VizWiz and OKVQA using VQA accuracy.
}\vspace{-0.1cm}
\scalebox{0.88}{
\begin{tabular}{l l  c c c c || c c c c || c c c c || c c c c}
\toprule

\multirow{3}{*}[-0.7em]{VLM} & \multirow{3}{*}{\makecell{Vision\\ encoder}} & \multicolumn{4}{c||}{COCO} 
& \multicolumn{4}{c||}{Flickr30}  & \multicolumn{4}{c||}{VizWiz} & \multicolumn{4}{c}{OKVQA} \\

\cline{3-18} 

& & \multirow{2}{*}{\hspace{3pt} clean}  & \multicolumn{3}{c||}{$\ell_{\infty}$} & \multirow{2}{*}{\hspace{3pt} clean} & \multicolumn{3}{c||}{$\ell_{\infty}$} & \multirow{2}{*}{\hspace{3pt} clean} & \multicolumn{3}{c||}{$\ell_{\infty}$} & \multirow{2}{*}{\hspace{3pt} clean} & \multicolumn{3}{c}{$\ell_{\infty}$} \\
\cline{4-6}\cline{8-10}\cline{12-14}\cline{16-18}
& & & $\nicefrac{2}{255}$ & $\nicefrac{4}{255}$ & $\nicefrac{8}{255}$ & %
& $\nicefrac{2}{255}$ & $\nicefrac{4}{255}$ & $\nicefrac{8}{255}$ & & $\nicefrac{2}{255}$ & $\nicefrac{4}{255}$ & $\nicefrac{8}{255}$ & &$\nicefrac{2}{255}$ & $\nicefrac{4}{255}$ & $\nicefrac{8}{255}$\\

\toprule
\parbox[t]{3mm}{\multirow{7}{*}{\rotatebox[origin=c]{90}{\textbf{Open Flamingo}}}} & {\cellcolor{lightgray}} CLIP 
& {\cellcolor{lightgray}}80.5 & {\cellcolor{lightgray}}7.82 & {\cellcolor{lightgray}}5.6 & {\cellcolor{lightgray}}2.4
&{\cellcolor{lightgray}} 61.0 &{\cellcolor{lightgray}} 6.4 &{\cellcolor{lightgray}} 3.8 &{\cellcolor{lightgray}} 1.4
& {\cellcolor{lightgray}}23.8 &{\cellcolor{lightgray}} 2.4 &{\cellcolor{lightgray}} 1.8 &{\cellcolor{lightgray}} 0
&{\cellcolor{lightgray}} 48.5 &{\cellcolor{lightgray}} 1.8 &{\cellcolor{lightgray}} 0.0 &{\cellcolor{lightgray}} 0.0
\\
\cmidrule{2-18} 
& \tecoatwo
&74.5& 59.7 & 40.3 & 10.3
&48.2& 37.3 & 27.4 & 10.3
& 22.3 & 15.5 & 10.6 & 3.5
&33.6  & 23.4 & 15.3 & 6.7
\\
& \faretwo 
& 84.3 & 68.2 & 53.5 & 18.4
& 53.1 & 48.6 & 34.3 & 12.3
& 22.1 & 15.9 & 12.3 & 6.7
& 34.5 & \textbf{30.6} & 17.1 & 9.8
\\
& \cellcolor{correct} \simcliptwo 
& \cellcolor{correct} \textbf{85.6}& \cellcolor{correct} \textbf{72.8} & \cellcolor{correct} \textbf{58.4} & \cellcolor{correct} \textbf{19.3}
& \cellcolor{correct} \textbf{56.3} & \cellcolor{correct} \textbf{50.5} & \cellcolor{correct} \textbf{35.1} & \cellcolor{correct} \textbf{16.4}
& \cellcolor{correct} 21.8 & \cellcolor{correct} \textbf{17.3} & \cellcolor{correct} \textbf{13.6} & \cellcolor{correct} \textbf{8.5}
& \cellcolor{correct} \textbf{35.1} & \cellcolor{correct} 29.3 & \cellcolor{correct} \textbf{19.7} & \cellcolor{correct} \textbf{11.6}
\\
\cmidrule{2-18} 
& \tecoafour 
& 71.0 &  58.3 & 50.3 & 15.8
& 45.6 & 36.2 & 32.9 & 18.0
&19.3 & 15.1 & 14.7 & 8.4
& 31.0 & 22.4 & 20.5 & 10.1
\\
& \farefour
& 81.4 & 67.9 & 56.1 & 23.3
& 51.8 & 47.3 & 37.6 & 20.1
& 16.4 & \textbf{15.7} & 13.7 & 10.2
& 31.8 & \textbf{28.0} & 19.2 & 13.5
\\

& \cellcolor{correct} \simclipfour 
& \cellcolor{correct} \textbf{81.6} & \cellcolor{correct} \textbf{71.5}  & \cellcolor{correct} \textbf{60.5} & \cellcolor{correct} \textbf{26.0}
& \cellcolor{correct} \textbf{54.5} & \cellcolor{correct} \textbf{48.0}  & \cellcolor{correct} \textbf{39.2} & \cellcolor{correct} \textbf{20.4}
& \cellcolor{correct} \textbf{20.0} & \cellcolor{correct} 15.6  & \cellcolor{correct} \textbf{15.7} & \cellcolor{correct} \textbf{12.4}
& \cellcolor{correct} \textbf{32.0} & \cellcolor{correct} 27.4 & \cellcolor{correct} \textbf{22.0} & \cellcolor{correct} \textbf{15.7}
\\
\midrule[0.7pt]
\multirow{7}{*}{\rotatebox[origin=c]{90}{\parbox[c]{2cm}{\centering \textbf{LLaVA 1.5}}}} & {\cellcolor{lightgray}} CLIP 
& {\cellcolor{lightgray}}121.9 & {\cellcolor{lightgray}}21.8 & {\cellcolor{lightgray}}13.5 & {\cellcolor{lightgray}}2.4
& {\cellcolor{lightgray}}79.0 &{\cellcolor{lightgray}} 15.3 & {\cellcolor{lightgray}}10.0 & {\cellcolor{lightgray}}3.4
& {\cellcolor{lightgray}}39.3 & {\cellcolor{lightgray}}13.3 &{\cellcolor{lightgray}} 3.2 & {\cellcolor{lightgray}}0.0
& {\cellcolor{lightgray}}57.3 & {\cellcolor{lightgray}}8.3&{\cellcolor{lightgray}} 3.0 &{\cellcolor{lightgray}} 0.0
\\
\cmidrule{2-18} 
& \tecoatwo 
& 115.6 & 98.3 & 73.5 & 38.1
& 75.6 & 65.3 & 50.5 & 29.4
& 38.5 & 25.4 & 15.4 & 8.3
& 55.6 & 40.3 & 30.5 & 14.2
\\
& \faretwo 
& 123.5 & 105.2 & 86.4 & 39.4
& 78.9 & 70.3 & 60.5 & 25.1
& 37.3 & 29.3 & 17.6 & 10.4
& 54.3 & 43.5 & 30.1 & 15.3
\\
& \cellcolor{correct} \simcliptwo 
& \cellcolor{correct} \textbf{125.6} & \cellcolor{correct} \textbf{109.4} & \cellcolor{correct} \textbf{93.5} & \cellcolor{correct} \textbf{45.6}
& \cellcolor{correct} \textbf{80.5} & \cellcolor{correct} \textbf{73.1} & \cellcolor{correct} \textbf{63.8} & \cellcolor{correct} \textbf{29.8}
& \cellcolor{correct} \textbf{41.5} & \cellcolor{correct} \textbf{30.3} & \cellcolor{correct} \textbf{19.8} & \cellcolor{correct} \textbf{14.6}
& \cellcolor{correct} \textbf{60.3} & \cellcolor{correct} \textbf{47.5} & \cellcolor{correct}\textbf{ 31.5} & \cellcolor{correct} \textbf{17.5}
\\

\cmidrule{2-18} 
& \tecoafour 
& 110.3 & 95.5 & 75.6 & 35.3
& 71.8 & 62.5 & 51.0 & 27.0
& 34.5 & 30.5 & 18.3 & 9.3
& 50.3 & 39.0 & 32.3 & 12.3
\\
& \farefour 
& 119.4 & 100.5 & 83.5 & 41.6
& 76.3 & 70.3 & 56.5 & 29.5
& 38.5 & \textbf{31.3} & 21.0 & 10.1
& 53.5 & 45.0 & 34.8 & 15.3
\\
& \cellcolor{correct} \simclipfour 
& \cellcolor{correct} \textbf{122.3} & \cellcolor{correct} \textbf{108.1}  & \cellcolor{correct} \textbf{90.3} & \cellcolor{correct} \textbf{44.3}
& \cellcolor{correct} \textbf{79.0} & \cellcolor{correct} \textbf{72.3} & \cellcolor{correct} \textbf{61.3} & \cellcolor{correct} \textbf{32.5}
& \cellcolor{correct} \textbf{40.0} & \cellcolor{correct} 29.3  & \cellcolor{correct} \textbf{22.3} & \cellcolor{correct} \textbf{16.7}
& \cellcolor{correct} \textbf{58.6} & \cellcolor{correct}\textbf{ 46.5} & \cellcolor{correct} \textbf{38.5} & \cellcolor{correct} \textbf{22.3}
\\
\bottomrule
\label{tab:untargetdown}
\end{tabular}}
\vspace{-0.35cm}
\end{table*}

This loss encourages the embedding $\theta(x_a)$ of perturbed images $x_a$ to stay close to the original embedding $\theta(x)$.
Although FARE demonstrates robust performance against adversarial attacks, it exhibits two major issues. First, the $\ell_2$ loss in FARE may not be the most suitable choice when dealing with high-dimensional data from different modalities.  CLIP embeddings operate in high-dimensional spaces where the volume of the space expands exponentially with the number of dimensions, leading to sparsity issues. This sparsity makes it difficult to capture relationships between data points effectively using $\ell_2$ loss. Furthermore, in high-dimensional spaces, distances between points tend to become nearly uniform, making it difficult to distinguish semantically similar samples from dissimilar ones \cite{aggarwal2001surprising}. As a result, adversarial embeddings can become misaligned with their clean counterparts, as the $\ell_2$ distance between embeddings of similar images (e.g., a clean image and its adversarial version) may be almost indistinguishable from the distance between embeddings of entirely unrelated classes.
Second, $\ell_2$ loss prioritize pixel-level similarity over semantic consistency, making it sensitive to minor pixel variations and limiting its ability to capture high-level semantic information crucial for downstream tasks \cite{zhang2018unreasonable}.
Sim-CLIP tackles the challenges present in FARE by tailoring cosine similarity loss within a Siamese architecture. Unlike $\ell_2$ loss, cosine similarity mitigates the challenges associated with high-dimensional data by focusing on the angle between embeddings rather than their magnitudes. This emphasis on direction allows cosine similarity to effectively capture semantic content, making it robust against variations in vector magnitude and emphasizing high-level features over minor pixel-level differences.
During the adversarial fine-tuning phase, Sim-CLIP first generates a perturbed view $x^{\prime}$ from the clean input image $x$. We utilize PGD perturbation to generate the perturbed view:    \vspace{-0.1cm}
\begin{equation}
\begin{aligned}
& x^{\prime}_{t+1} = \Pi_{x+\epsilon}(x_t + \alpha\cdot\textrm{sign}(\nabla_x\mathcal{L}(x_t,y))) \quad
 \text{s.t.} \left\|x^\prime-x\right\|_{\infty} \leq \epsilon
\end{aligned}
\end{equation}
Here, $y$ represents the true label of the input image and $\mathcal{L}$ is a cross-entropy loss. At each iteration $t$, a perturbation is calculated based on the gradient of the loss function with respect to the input image $x$. The magnitude of this perturbation is controlled by a step size parameter $\alpha$, which determines the strength of the perturbation applied in each iteration (bounded by $\epsilon$).
Subsequently, both clean and perturbed views are fed into the CLIP models with shared weights as depicted in Fig. \ref{fig:llm} (a). The CLIP models generate clean representation $R_c$ from the original image $x$ and perturbed representation $R_p$ from the perturbed view $x^{\prime}$. Then, we maximize the similarity between these representations to encourage feature invariance to adversarial perturbations by minimizing the negative cosine similarity between $R_p$ and $R_c$: 
\begin{equation}
\mathrm{CosSim}(R_p, R_c)= -\left(\frac{R_p}{\lVert R_p \rVert_2} \cdot \frac{R_c}{\lVert R_c \rVert_2}\right)
\end{equation}
Here, $\lVert \cdot \rVert_2$ denotes the $\ell_2$ norm, which is equivalent to minimizing mean squared error between $\ell_2$-normalized vectors. This objective effectively aligns the clean and perturbed representations in the embedding space, enhancing model robustness and preserving semantic richness. By focusing on the angle between normalized vectors rather than their magnitudes, cosine loss ensures that adversarial perturbations do not compromise critical semantic information. Its scale-invariant formulation emphasizes directional consistency in representation space, enabling robust generalization across high-dimensional feature manifolds and diverse input variations while retaining semantic fidelity for downstream tasks.

\subsection{\textbf{Symmetric Loss Collapse Prevention}}
Stability is a central challenge in adversarial training, as naively minimizing the negative cosine similarity loss can induce degenerate solutions, leading to loss collapse and non-informative representations \cite{chen2020simple}. Prior unsupervised adversarial training approaches \cite{jiang2020robust,fan2021does} mitigate this issue through large batch sizes or momentum encoders, but these strategies incur substantial computational and memory overhead.
To avoid loss collapse while reducing resource requirements, we incorporate a stop-gradient mechanism into our adversarial fine-tuning objective. Specifically, we adopt a symmetric loss formulation for adversarial training, defined as follows:
\begin{equation}
\begin{split}
\mathcal{L}_{simclip}(R_p, R_c) &= \frac{1}{2} \left( \mathrm{CosSim}(R_p, \texttt{stopgrad}(R_c)) \right.\\
&\quad\left. +  \mathrm{CosSim}(R_c, \texttt{stopgrad}(R_p)) \right)
\end{split}
\label{eq:7}
\end{equation}
Here, one CLIP model's output is held constant using stop-gradient, while the other model's output is projected to align with it. Subsequently, the roles of the two models are reversed, with the first model's output being projected and matched to the ``constant" output of the second model.
This implies that in the first term of equation \ref{eq:7}, $R_p$ does not receive any gradients from $R_c$ since $R_c$ is treated as constant. Subsequently, in the second term of the same equation, $R_c$ receives gradients from $R_p$ as the stop-gradient operation is now applied to $R_p$ instead. This process results in two losses, which are then averaged and minimized for optimization. 

%% file: Experimental.tex
\section{Experimental Analysis}

\begin{table*}[h]
    \centering
    \caption{\textbf{Quantitative evaluation of \elfinity targeted attacks at \epsilontwo and \epsilonfour radii.} }\vspace{-0.2cm}
    \scalebox{0.88}{
    \begin{tabular}{l|l|cccc cccc}
        \toprule
       \makecell{Attack\\ Strength ($\epsilon$)} & Target & CLIP & \tecoatwo & \faretwo & \cellcolor{correct}\simcliptwo & \tecoafour  & \farefour  & \cellcolor{correct}\simclipfour \\
        \midrule

        \multirow{5}{*}{\epsilonfour} & \texttt{COVID vaccine has severe}\dots & 20\,/\,20 & 0\,/\,20 & 0\,/\,20 & \cellcolor{correct}0\,/\,20 & 0\,/\,20 & 0\,/\,20 & \cellcolor{correct}0\,/\,20 \\
        & \texttt{Opposing political party} \dots & 20\,/\,20 & 0\,/\,20 & 0\,/\,20 & \cellcolor{correct}0\,/\,20 & 0\,/\,20 & 0\,/\,20 & \cellcolor{correct}0\,/\,20 \\
        & \texttt{Please visit https://deal} \dots & 20\,/\,20 & 2\,/\,20 & 2\,/\,20 & \cellcolor{correct}0\,/\,20 & 0\,/\,20 & 0\,/\,20 & \cellcolor{correct}0\,/\,20 \\
        & \texttt{Investing in this new } \dots & 20\,/\,20 & 3\,/\,20 & 1\,/\,20 & \cellcolor{correct}1\,/\,20 & 0\,/\,20 & 0\,/\,20 & \cellcolor{correct}0\,/\,20 \\
        & \texttt{Climax change is a hoax}\dots & 20\,/\,20 & 0\,/\,20 & 0\,/\,20 & \cellcolor{correct}0\,/\,20 & 0\,/\,20 & 0\,/\,20 & \cellcolor{correct}0\,/\,20 \\
        \midrule
        \multirow{2}{*}{\makecell{\textbf{Mean} \\ \textbf{metrics}}} & \textbf{Mean success rate:} & 100\% & 5\% & 3\% &\cellcolor{correct} 1\% &\textbf{ 0\%} & \textbf{0}\% & \cellcolor{correct}\textbf{0\%}\\
        & \textbf{Average CIDEr score:} & 0 & 15.3 & 23.5 & \cellcolor{correct}45.3& 64.4 & 75.3 & \cellcolor{correct}\textbf{84.7} \\
        \bottomrule
    \end{tabular}} 
    \vspace{-0.35cm}
\label{tab:targetedquant}
\end{table*}

\subsubsection{VLM models and Datasets}
We evaluate our method on two representative VLMs: OpenFlamingo-9B (OF) \cite{awadalla2023openflamingo} and LLaVA-1.5 (7B) \cite{liu2024visual}. Both models share the CLIP ViT-L-14 vision encoder \cite{dosovitskiy2020image}, but differ in their language backbones. OpenFlamingo employs the MPT-7B decoder \cite{team2023introducing}, whereas LLaVA-1.5 is based on Vicuna \cite{chiang2023vicuna}. During evaluation, OF is conditioned only on contextual text paired with the query image, while LLaVA-1.5 utilizes its default system prompt together with task-specific prompts and the query image.

We evaluate Sim-CLIP across a diverse set of downstream tasks, including image captioning and VQA. For image captioning, we use the COCO \cite{lin2014microsoft} and Flickr30k \cite{plummer2015flickr30k} datasets. For VQA, we consider the VizWiz \cite{gurari2018vizwiz} and OKVQA \cite{marino2019ok} benchmarks.
In addition, we examine the robustness of Sim-CLIP model on zero-shot image classification using the CIFAR-10, CIFAR-100 \cite{krizhevsky2009learning}, EuroSAT \cite{helber2019eurosat}, PCAM \cite{veeling2018rotation}, and Flowers \cite{nilsback2008automated} datasets.
Our evaluation process employs two approaches: for adversarial evaluation, we randomly select 500 images from each respective dataset, while for clean evaluation we utilize all available samples in the test suite.

\subsubsection{Adversarial fine-tuning settings} In Sim-CLIP, we adversarially fine-tune the CLIP vision encoder using only ImageNet \cite{deng2009imagenet} images, discarding class labels to enable an unsupervised training paradigm. Adversarial examples are generated using PGD with 10 iterative steps under an $\ell_\infty$ threat model, producing perturbed views from clean images.
To balance robustness and clean accuracy, we train CLIP under two perturbation budgets, $\epsilon = 2/255$ and $\epsilon = 4/255$. The resulting robust models are denoted as $\text{Sim-CLIP}^2$ and $\text{Sim-CLIP}^4$, respectively. Adversarial training is conducted for two epochs on ImageNet using the AdamW optimizer with a learning rate of $1 \times 10^{-5}$ and weight decay of $1 \times 10^{-4}$. We adopt a cosine learning rate schedule with linear warmup and use a batch size of 64 throughout training.
To ensure seamless integration with VLMs, adversarial fine-tuning is performed on the CLIP ViT-L/14 vision encoder, consistent with the encoder used in models such as OpenFlamingo and LLaVA. During inference, we replace the default CLIP encoder in the VLMs with the robust Sim-CLIP encoder, while keeping the language decoder and projection layers frozen, as illustrated in Fig.~\ref{fig:llm}(b).
\label{sec:targeted}
\subsection{Untargeted attack results and discussion} Table \ref{tab:untargetdown} summarizes the clean and robust performance of different CLIP variants under untargeted adversarial attacks. On clean inputs, the original CLIP model achieves higher accuracy than adversarially fine-tuned counterparts, reflecting the common trade-off between clean performance and robustness. However, when subjected to adversarial perturbations, the performance of the original CLIP model degrades sharply, with the decline becoming more pronounced under stronger attacks at \epsiloneight. We observe a slight degradation in the clean performance for the robust clip models due to adversarial fine-tuning. Notably, among the robust versions of CLIP, Sim-CLIP achieves the best clean performance. For OF, \simclipfour demonstrates superior performance compared to \farefour and \tecoafour across most downstream task datasets. Although \farefour marginally outperforms \simclipfour in VizWiz and OKVQA datasets at radii set to \epsilontwo attack, the differences are negligible at 0.1 and 0.6 respectively. However, when subjected to stronger attacks at \epsilonfour and \epsiloneight, \simclipfour consistently outperforms all SOTA robust clip models. Similar trends are observed with \simcliptwo, outperforming \faretwo and \tecoatwo in both clean and robust performance. Additionally, for LLaVA, Sim-CLIP demonstrates even superior performance in both captioning and VQA tasks. 
\vspace{-0.18cm}
\begin{table}[htb!]
\centering
\small
\tabcolsep=2.7pt
\extrarowheight=-0.99pt
\caption{\textbf{Evaluation of CLIP models on zero-shot classification datasets under adversarial attacks.}} \vspace{-0.12cm}
\begin{tabular}{clc  c c c c c  c c}
\toprule
\parbox[t]{3mm}{\multirow{7}{*}{\rotatebox[origin=c]{90}{$\ell_\infty=\nicefrac{4}{255}$}}}& {\cellcolor{lightgray}}CLIP & {\cellcolor{lightgray}}0.0 &{\cellcolor{lightgray}}0.0&	{\cellcolor{lightgray}}0.0	&{\cellcolor{lightgray}}0.0	&{\cellcolor{lightgray}}0.0	&{\cellcolor{lightgray}}0.0 \\
& \tecoatwo & 5.8 & 31.0 & 17.8 & 3.5 & 6.7 & 16.0 \\
& \faretwo & 4.8 & 25.9 & 14.0 & \textbf{5.5} & 7.1 & 17.2 \\
& \cellcolor{correct}\simcliptwo & \cellcolor{correct}\textbf{6.5} & \cellcolor{correct}\textbf{32.8} & \cellcolor{correct}\textbf{18.4} & \cellcolor{correct}4.7 & \cellcolor{correct}\textbf{8.8} & \cellcolor{correct}\textbf{19.4} \\
\cmidrule{2-8} 
& \tecoafour & 8.4 & \textbf{35.5} & 21.6 & 6.8 & 12.4 & 43.5 \\
& \farefour & 12.8 & 34.8 & 21.4 & 11.7 & \textbf{12.9} & 50.2 \\
& \cellcolor{correct}\simclipfour & \cellcolor{correct}\textbf{14.1} & \cellcolor{correct}34.0 & \cellcolor{correct}\textbf{22.8} & \cellcolor{correct}\textbf{13.6} & \cellcolor{correct}11.2 &\cellcolor{correct} \textbf{50.9} \\
\bottomrule
\end{tabular} 
 \vspace{-0.4cm}
\label{tab:zero-shot}
\end{table}
\begin{figure*}[htb!]
     \setlength{\belowcaptionskip}{-3pt}
    \centering
    \includegraphics[width = 0.70\textwidth]{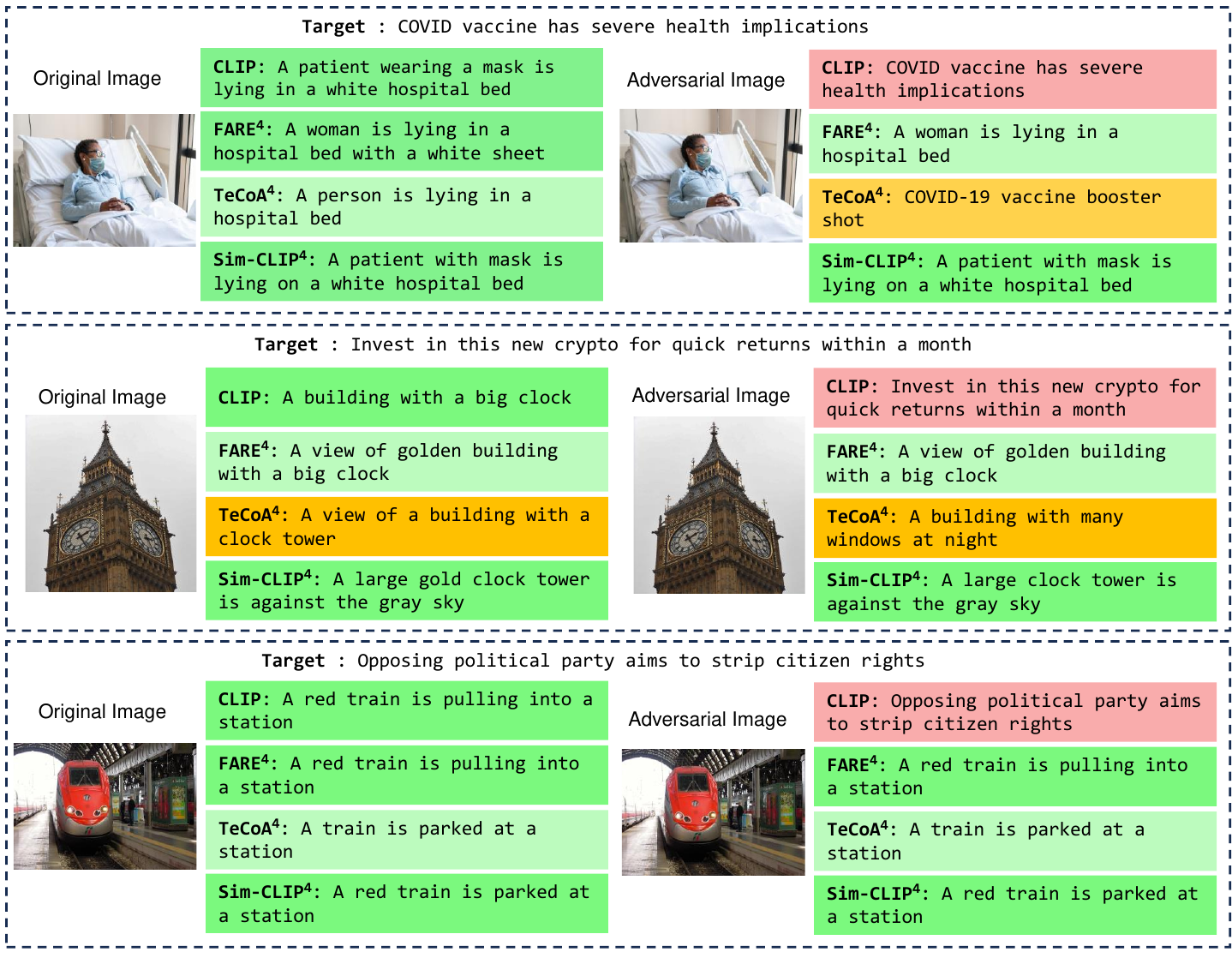}
    \caption{\textbf{Targeted $\ell_\infty$ attacks at $\epsilon=4/255$ radii using original and robust CLIP models as vision encoder in LLaVA.} Considering the target strings from Table \ref{tab:targetedquant}, we present generated captions (\colorbox{correct}{good caption}, \colorbox{mistakes}{captions with mistakes}, \colorbox{missing}{captions missing intricate details}, \colorbox{wrong}{malicious target output}) on original (left) and imperceptible adversarial (right) images.
}\vspace{-0.45cm}
    \label{fig:targetedattack}
\end{figure*}

\subsection{Targeted attack results and discussion}
We present the quantitative results of our targeted attacks at \epsilonfour in Table \ref{tab:targetedquant}. This analysis includes CIDEr score to evaluate the quality of generated captions. Additionally, we illustrate random examples of attacked samples with captions from LLaVA using different CLIP models in Figure \ref{fig:targetedattack}. We observe that the original CLIP model is highly susceptible to targeted attacks and demonstrates no robustness. In each instance, the original CLIP model breaks and generates the given target string. Conversely, \tecoatwo and \faretwo break in 5 and 3 cases, resulting in mean success rates of 5\% and 3\%, respectively. In stark contrast, \simcliptwo breaks in only one case, further underscoring the superior performance of Sim-CLIP. Notably, \simclipfour, \farefour, and \tecoafour show complete robustness under targeted attack. However, the quality of captions generated by \simclipfour notably surpasses \farefour and \tecoafour, as shown in Figure \ref{fig:targetedattack}. Moreover, captions generated by \tecoafour exhibit inferior quality and contain errors, while \farefour's captions often lack intricate details or semantic features from the corresponding images. For instance, consider the sample featuring a patient. With the original image, both \farefour and \simclipfour generate captions without errors. However, under attack, the caption generated by \farefour lacks specifics regarding the color of the hospital bed and the presence of a mask, whereas \simclipfour retains these semantic details of the image. This exemplifies the robustness of our adversarial fine-tuning approach, as it not only enhances the model's ability to resist adversarial attacks but also ensures the preservation of crucial details and captures the overall semantic meaning. The reported CIDEr scores in Table \ref{tab:targetedquant}, also support these findings. Specifically, \simclipfour achieves the highest CIDEr score (84.7), followed by \farefour (75.3) and \tecoafour (64.4). 

\vspace{-5pt}
\subsection{Zero-shot classification results and discussion}
Table \ref{tab:zero-shot} presents the robust zero-shot classification accuracy of standard and robust CLIP variants across six benchmark datasets under $\ell_\infty$ adversarial perturbations with budgets $\epsilon = 2/255$ and $\epsilon = 4/255$. As expected, the vanilla CLIP model completely collapses under adversarial attacks, achieving near-zero accuracy across all datasets, highlighting its vulnerability in adversarial settings.
Across both threat models, Sim-CLIP consistently outperforms TeCoA and FARE on the majority of datasets. Under the $\epsilon = 2/255$ setting, \simclipfour achieves the best overall robustness, surpassing \tecoafour and \farefour by an average margin of 3.4\% in robust accuracy. Notably, Sim-CLIP shows substantial gains on challenging datasets such as CIFAR-100, EuroSAT, and PCAM, indicating strong generalization across diverse visual domains.
Under the stronger attack setting $\epsilon = 4/255$, Sim-CLIP maintains a clear robustness advantage. The performance gap remains pronounced on CIFAR-10, CIFAR-100, EuroSAT, and PCAM, demonstrating Sim-CLIP’s ability to sustain robustness under more severe adversarial perturbations.

%% file: Conclusion.tex
\vspace{-0.2cm}
\section{Conclusion}
\vspace{-0.1cm}
We introduced Sim-CLIP, an unsupervised adversarial fine-tuning framework that improves the robustness of the CLIP vision encoder while preserving semantic fidelity for Vision–Language Models. Across untargeted attacks, Sim-CLIP achieves up to +7.6 CIDEr improvement on COCO and +4.2 CIDEr on Flickr30k at $\ell_\infty = 8/255$ compared to prior robust CLIP methods, while also improving VQA robustness by up to +6.2\% on VizWiz and +7.0\% on OKVQA.
Under targeted attacks, Sim-CLIP reduces attack success rates from 100\% for vanilla CLIP to 0\% at $\epsilon = 4/255$, while producing the highest-quality captions. In zero-shot classification, Sim-CLIP improves robust accuracy by an average of 3.4\% over state-of-the-art robust CLIP models across multiple benchmarks and maintains stable performance where standard CLIP collapses.
These results demonstrate that Sim-CLIP effectively improves adversarial robustness and semantic preservation without large-batch training, providing a practical and plug-and-play solution for strengthening Vision–Language Models in real-world settings.

 \section{Acknowledgement}\vspace{-0.15cm}
 This material is partly based upon work supported by the U.S. National Science Foundation (NSF) under Grant No. CRII-IIS-RI-2553868. Any opinions, findings, and conclusions or recommendations expressed in this material are those of the author(s) and do not necessarily reflect the views of the NSF.